\documentclass{article}
\usepackage[a4paper]{geometry}
\usepackage[utf8]{inputenc}
\usepackage[T1]{fontenc}
\usepackage{graphicx}
\usepackage{amsmath,amssymb,amsfonts}
\usepackage{tabularx,booktabs}
\usepackage{xcolor} 
\definecolor{colorkb}{HTML}{F8474A}
\usepackage[colorlinks=true,
            linkcolor=colorkb,
            citecolor=colorkb,
            urlcolor=colorkb]{hyperref}
\usepackage{xurl}
\usepackage{xcolor}
\usepackage[numbers]{natbib}
\usepackage{subcaption}

\title{\textbf{Khana: A Comprehensive Indian Cuisine Dataset}}
\author{
  Omkar Prabhu\\
  \texttt{prabhuomkar@pm.me}
}
\date{}
\begin{document}

\maketitle

\begin{abstract}

As global interest in diverse culinary experiences grows, food image models are essential for improving food-related applications by enabling accurate food recognition, recipe suggestions, dietary tracking, and automated meal planning. Despite the abundance of food datasets, a noticeable gap remains in capturing the nuances of Indian cuisine due to its vast regional diversity, complex preparations, and the lack of comprehensive labeled datasets that cover its full breadth. Through this exploration, we uncover \texttt{Khana}, a new benchmark dataset for food image classification, segmentation, and retrieval of dishes from Indian cuisine. Khana fills the gap by establishing a taxonomy of Indian cuisine and offering around \texttt{131K} images in the dataset spread across \texttt{80} labels, each with a resolution of \texttt{500x500} pixels. This paper describes the dataset creation process and evaluates state-of-the-art models on classification, segmentation, and retrieval as baselines. Khana bridges the gap between research and development by providing a comprehensive and challenging benchmark for researchers while also serving as a valuable resource for developers creating real-world applications that leverage the rich tapestry of Indian cuisine. \\

\begin{center}
\textbf{Webpage:} \url{https://khana.omkar.xyz}
\end{center}

\end{abstract}

\begin{figure*}
  \centering
  \includegraphics[width=0.75\linewidth]{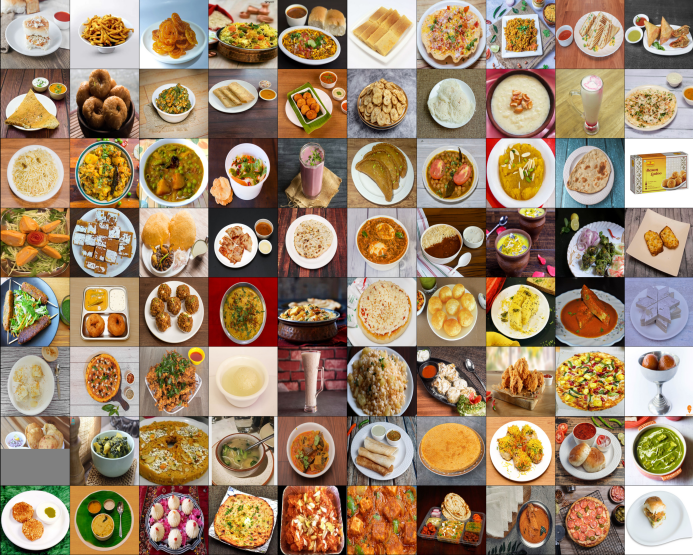}
	\caption{Representative images from each category}
	\label{fig:heading}
\end{figure*}

\section{Introduction}
\label{sec:introduction}

In this digital age, food transcends the physical realm due to the exponential rise of smartphones and social media with vibrant pictures. With this rise, there is a need for efficient navigation across diverse culinary landscapes for several delivery and food review platforms, as well as for precise dietary assessment and personalized nutrition. Food image classification and food retrieval plays a critical role here, but the journey from pixel to palate presents unique challenges for all cuisines. \cite{arxiv.1907.06167, Min2020, Min2023} impacted food classification by establishing benchmark datasets, pioneering architectural advancements, and showcasing practical implementations. They have significantly enriched the field by introducing pivotal datasets and innovative architectures while highlighting various challenges encountered in fine-grained categorization throughout the years.

Indian cuisine, in particular, is a kaleidoscope of flavors and textures. Different spices, textures, and ways of cooking create several delicious varieties of food. The close resemblance often masks these unique flavors and details. This heterogeneity becomes a significant hurdle for image classification algorithms trained on broader datasets. The need for detailed categorization in a convoluted domain like food emphasizes the importance of fine-grained image classification, which involves recognizing subtle visual nuances. Food classification has seen considerable effort, especially in Western and Asian cuisines. Even in the case of Asian cuisines, the emphasis has predominantly been on Japanese or Chinese influences. The research conducted thus far has significantly contributed to the advancement of food classification within these specific culinary domains.

Recognizing this gap, we introduce Khana, a comprehensive benchmark dataset for fine-grained food image classification in the Indian context, as the main contribution. It contains \texttt{131K+} images with \texttt{80} categories belonging to different super-classes, such as breakfast, main course, snacks, and beverages. Beyond its sheer size and curated composition, Khana paves the way for a future where the nuances of Indian food can be readily interpreted by machines, bridging the gap between the visual and the delectable. In addition, we provide taxonomy with extensive experiments comparing various state-of-the-art methods of image classification. We hope to empower research, fuel innovation, and celebrate the diversity and richness of Indian food, one pixel at a time.

\section{Related Work}
\label{sec:related}

This section highlights prior noteworthy research focusing on food-related datasets, work on food classification, segmentation and retrieval.

\subsection{Food Related Datasets}
\label{subsec:related-data}

Food datasets as shown in Table \ref{tab:comparison-datasets} have grown in recent years, covering different cuisines, cooking styles, and annotation formats developed for food-related tasks like classification, segmentation, and retrieval. Most of these datasets consist of Eastern and Western food dishes, with very little (mostly fine-grained) work done on cuisines like Chinese \cite{arxiv.1705.02743}, Japanese \cite{Jianing2019}, Brazilian \cite{arxiv.2012.03087}, Kenyan \cite{Jalal2019}, and Singapore \cite{Sahoo2019}.  Despite the global prevalence and increasing popularity of Indian cuisine \cite{tasteatlas}, it remains underrepresented in food-related research work. The existing research lacks representation of Western dishes adapted to Indian tastes, distinctive regional cooking techniques, and the vibrant cultural elements that characterize Indian cuisine. Although the datasets generated until now are either scraped from the web or captured using manual labor, they do not tap into the widespread network of social media and food delivery applications. While the existing research is valuable, it lacks coverage of Indian cuisine, creating a need for exploratory analysis and benchmarking that would benefit the research community.

\begin{table*}
  \centering
  \renewcommand{\arraystretch}{1.25}
  \begin{tabular}{lllll}
    \hline
    \textbf{Dataset} & \textbf{Images} & \textbf{Categories} & \textbf{Cuisine} & \textbf{Year} \\
    \hline
    ChineseFoodNet \nocite{arxiv.1705.02743} & 180K & 208 & Chinese & 2017 \\
    THFOOD-50 \nocite{Termritthikun2017} & 15K & 50 & Thai & 2017 \\
    Food524DB \nocite{Ciocca2017} & 247K & 524 & Misc. & 2017 \\
    Food-101N \nocite{Lee2018} & 310K & 101 & Misc. & 2018 \\
    KenyanFood13 \nocite{Jalal2019} & 8K & 13 & Kenyan & 2019 \\
    FoodX-251 \nocite{arxiv.1907.06167} & 158K & 251 & Misc. & 2019 \\
    SUEC Food \nocite{arxiv.1903.07437} & 32K & 256 & Asian & 2019 \\
    Sushi-50 \nocite{Jianing2019} & 3.9K & 50 & Japanese & 2019 \\
    FoodAI-756 \nocite{Sahoo2019} & 400K & 756 & Singapore & 2019 \\
    ISIA Food-500 \nocite{Min2020} & 399K & 500 & Misc. & 2020 \\
    MyFood \nocite{arxiv.2012.03087} & 1250 & 9 & Brazilian & 2020 \\
    FoodSeg154 \nocite{Wu2021} & 10K & 154 & Misc. & 2021 \\
    Food2K \nocite{Min2023} & 1M & 2000 & Misc. & 2023 \\ 
    DailyFood-172 \nocite{Liu2024} & 42K & 172 & Misc. & 2024 \\
    AI4Food-NutritionDB \nocite{RomeroTapiador2024} & 558K & 893 & Misc. & 2024 \\\hline
    \textbf{Khana} & \textbf{130K} & \textbf{80} & \textbf{Indian} & \textbf{2025} \\\hline
  \end{tabular}
  \caption{Comparison of the Khana dataset with prior related works}
  \label{tab:comparison-datasets}
\end{table*}

\subsection{Food Classification and Segmentation}
\label{subsec:related-class-seg}

Food classification and segmentation are becoming more prominent in recent years, with real-world applications like dietary assessment, automated food logging, and nutritional analysis. Foundational work from \cite{Bossard2014} established a benchmark with 101 food categories that has driven later research in automated food recognition. There were early deep learning experiments \cite{arxiv.1606.05675} that demonstrated the benefits of CNNs for dietary assessments. We see a gradual evolution of fine-grained classification with regional cuisines like \cite{arxiv.1705.02743}, \cite{arxiv.1907.06167}, and transfer learning like \cite{Lee2018}. We also see large-scale neural networks for food recognition in \cite{Min2020}, \cite{Min2023}, and \cite{Liu2024}. For faster latency mobile use cases, there was a novel idea in \cite{Termritthikun2017}. The transition from classification to segmentation happened via \cite{Wu2021}, which enabled pixel-level food understanding. Contemporary research for real-world deployment challenges was presented in \cite{Liu2024}, and nutrition-focused recognition systems with \cite{Doyen2019} and \cite{RomeroTapiador2024}. Along with all these, \cite{Liu2025} highlights the transition from traditional computer vision to sophisticated deep learning architectures for food classification and segmentation tasks. 

\subsection{Food Retrieval}
\label{subsec:related-retrieval}

Food retrieval has become more prevalent in recent years due to its useful applications, such as dietary management, recipe recommendation, restaurant services, and food logging. \cite{arxiv.2003.03955} uses bi-directional LSTMS for encoding recipes and images in a common embedding space, but it struggles with noisy images and cross-domain matching between images and texts. \cite{Ciocca2017} uses CNNs for food retrieval based on embeddings with distance metrics, and it points out challenges due to noisy and cluttered food images, large intra-class variations due to differences in preparation, portion size, and presentation, as well as high inter-class similarity among visually overlapping dishes. \cite{arxiv.1810.06553} proposes bi-directional retrieval using CNNs for visual and deep language models to process ingredients and instructions. It suffers from a gap between visual and textual representations, variability in food presentation and image quality, and incomplete or ambiguous recipe descriptions. \cite{arxiv.1905.01273} proposes adversarial training for CNNs to learn cross-modal embeddings that map food images with recipe text for retrieval. It addresses challenges like domain shift between visual and textual modalities, noisy and ambiguous recipe descriptions, and the high visual similarity of different dishes. \cite{Shimoda2017} focuses on using CNNs to extract visual features and learning a similarity metric to improve retrieval performance. \cite{Min2020}, designed to advance recognition and retrieval tasks, uses global attention to capture holistic dish appearance and local attention to highlight discriminative regions such as textures or ingredients, which helps in large-scale settings. Most of the studies highlight challenges such as large intra-class variation due to different preparation and presentation styles, high inter-class similarity among visually related dishes, and noisy image conditions that make retrieval difficult.

\section{Khana Dataset}
\label{sec:data}

Given the considerable progress in food domain, existing datasets exhibit a clear lack of diversity. In particular, Indian cuisine, that is one of the largest and most varied culinary traditions remains underrepresented. To overcome this limitation, we present the research community with a novel single modality self-collected dataset and arranged using smart taxonomoy.

\subsection{Establishing a Taxonomy for Indian Cuisine}
\label{subsec:data-taxo}

The purpose of the taxonomy is two-fold: First, it aims to organize and structure the food items by establishing hierarchical relationships and culinary categories based on their preparation methods, regional origins, and cultural significance. Second, the taxonomy provides well-defined categories and subcategories that serve as training labels for usability in tasks like classification, segmentation, and retrieval. The taxonomy facilitates semantic search capabilities by establishing relationships between similar food items and cooking techniques.

\begin{figure}
	\centering
	\includegraphics[width=0.75\linewidth]{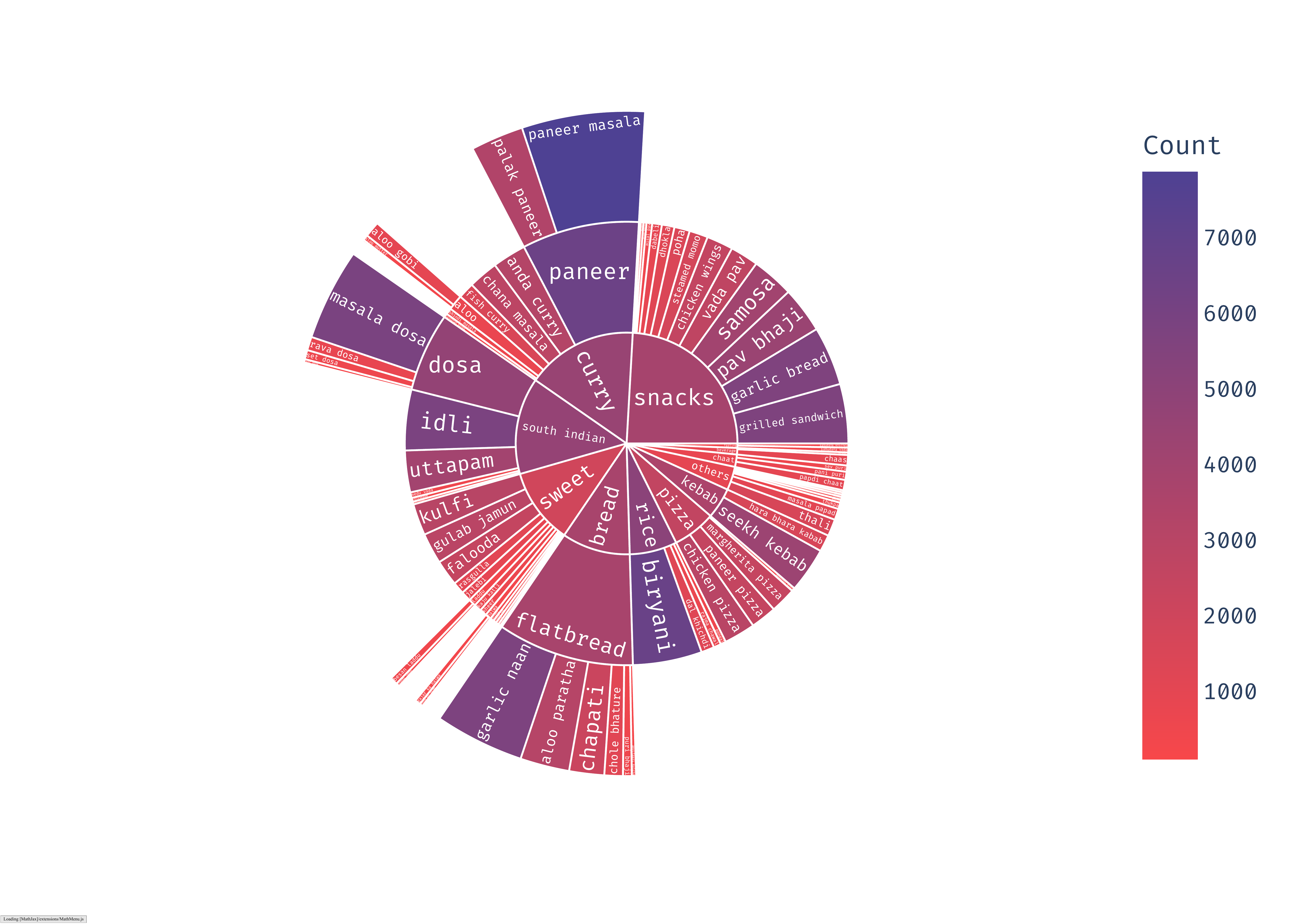}
	\caption{Hierarchical sunburst showing categories, dishes, varieties, and image counts per dish variety}
	\label{fig:sunburst}
\end{figure}

Figure \ref{fig:sunburst} shows the hierarchy and the number of images per dish variety. Each label is a dish variety in the dataset that belongs to a food category i.e. $\texttt{category} \rightarrow \texttt{dish} \rightarrow \texttt{variety}$. Dishes are categorized based on their ingredients, cooking methods, and regional cuisine. For example, \textit{dosa}, \textit{idli}, \textit{uttapam}, \textit{medu vada} are all dishes made up of ingredients like lentils and rice categorized as \textit{south indian} based on their regional nature. For example, \textit{anda curry}, \textit{chana masala}, \textit{fish curry}, \textit{bhindi masala} are all dishes categorized as \textit{curry} based on their cooking method and food consistency. Pav and chutney emerge as integral components, contributing to the dataset’s diversity. \textit{pav}, a type of bread roll, accompanies several food dishes across India in different forms. The dataset unfolds a fascinating narrative of how one dish can manifest based on its geographical origin. \textit{aloo}-related, \textit{puri}-related, and \textit{bread}-centered dishes exhibit distinct regional adaptations, reflecting the cultural nuances and local preferences as famous in different parts of India. For instance, \textit{aloo}-related dishes may vary in preparation and spice profiles, with regional influences shaping the culinary identity of each sample.

The taxonomy is built with flexibility and scalability at its core, where it can expand as culinary traditions evolve or as specific focus areas gain prominence. The categories and dishes can grow without disrupting the existing hierarchy. For example, \textit{curry} category can be expanded to different dishes based on different regions. Food item innovations in urban areas can be expanded into \textit{snacks} category. It also addresses cross-cultural and fusion dishes through flexible categorization, such as \textit{pizza} with Indian adaptations \textit{paneer pizza}. It also has traditional varieties and dishes that span different categories like \textit{steamed momo} and \textit{seekh kebab} tagged as both veg and non-veg under dietary preferences to reflect their nature.

\subsection{Data Collection and Cataloguing Addressing Multilingual Conventions}
\label{subsec:data-collect-catal}

The dataset showcases rich culinary diversity, spanning appetizers, main courses, desserts, snacks, and beverages. It features iconic dishes like \textit{dosa}, \textit{biryani}, \textit{gulab jamun}, and \textit{chaas}, with a wide variety of dishes such as \textit{flatbreads}, \textit{curries}, and \textit{sweets}. Both vegetarian and non-vegetarian options, including \textit{paneer} dishes and \textit{fish curry}, are well-represented, highlighting regional flavors and dietary preferences. Special categories like \textit{fasting foods} and \textit{kebabs} further enrich the taxonomy, reflecting the breadth of Indian and global cuisine.

The data for the Khana dataset comes from search engines and online food delivery platforms like Swiggy and Zomato \cite{swimto}. Web crawlers gathered images in an automated manner from keyword search results and restaurant delivery menu lists. The process removed duplicate images across restaurant menus and search results for varieties of the same dish by finding nearest neighbors using image embeddings generated from torchvision models \cite{torchvision}. Simple scripts filtered out low-quality images that did not meet the resolution criteria.

The images in the dataset are structured as per the dish variety names and there is a taxonomy CSV which contains information like \texttt{category}, \texttt{dish}, \texttt{variety}, \texttt{dietary} for each label in the dataset. There is no additional information such as preparation method or timestamps available.
 
We labeled the images using an automated approach during the collection stage, based on keyword searches or substring matches with restaurant menu item names. To ensure consistent labeling, we grouped samples containing varied Hinglish keywords for the same dish variety. For example, the label \textit{pani puri} can be also denoted as \textit{pani poori}, \textit{golgappa}, \textit{panipuri} or \textit{panipoori}. We filtered out images that contained a combination of different labels. To ensure label accuracy, the image folder for each label was manually verified by three annotators, who achieved inter-annotator agreement for classifying certain samples. For example, lot of \textit{south indian} dishes were available as a single combo-item in the dataset, which needed to be filtering.

There were no image processing techniques such as rotation, flipping or color adjustments used on the dataset to ensure its originality. There are no augmented samples in the dataset.

\subsection{Dataset Statistics and Characteristics}
\label{subsec:data-stats}

\begin{figure}[h]
	\centering
	\includegraphics[width=\linewidth]{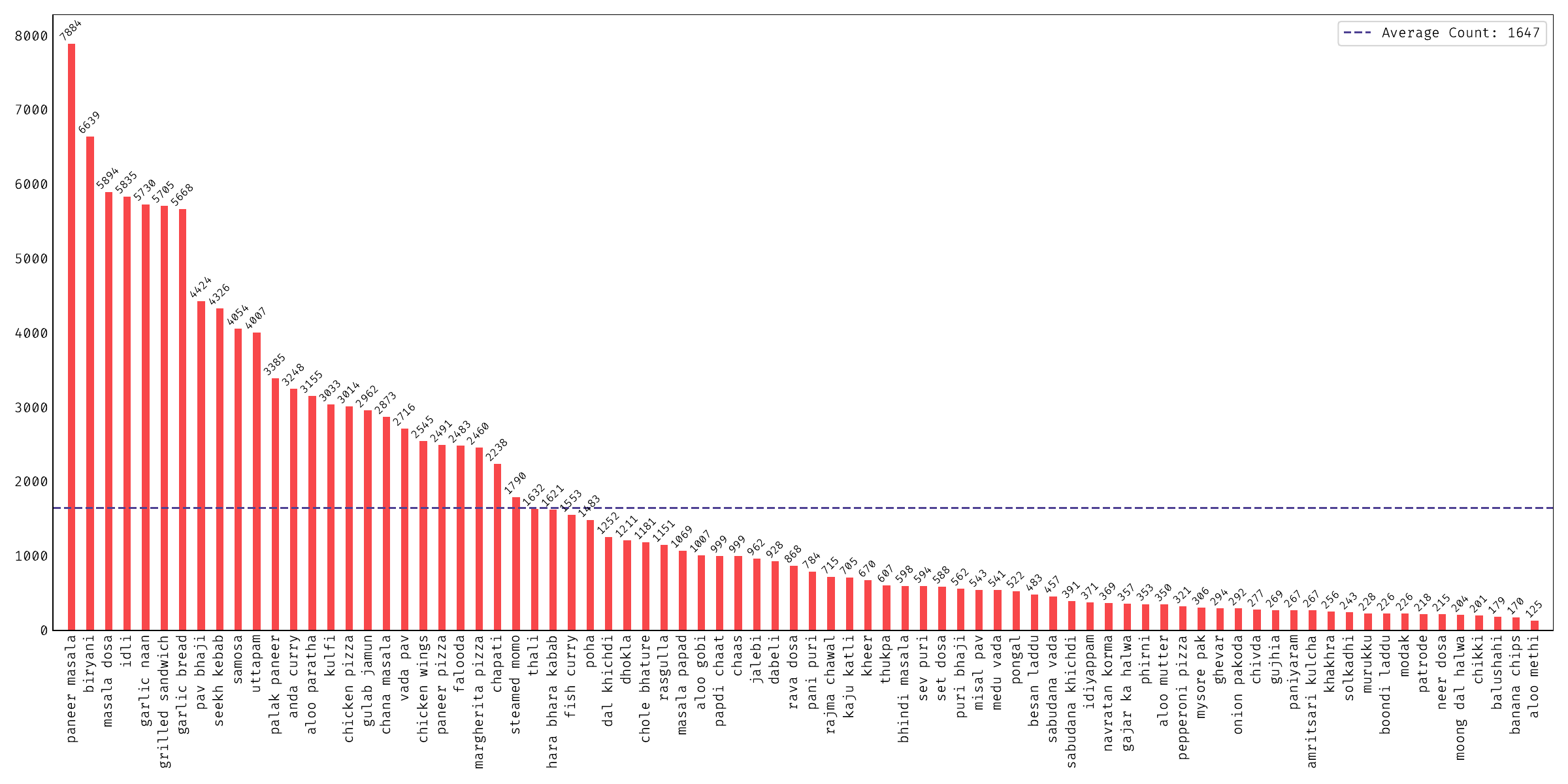}
	\caption{The distributions over each category}
	\label{fig:statistics}
\end{figure}

The dataset contains around \texttt{131K} images spread across \texttt{80} different classes. Each sample in the datset has a resolution of \texttt{500x500} pixels. The dataset is split into training, validation, and test sets with \texttt{70\%} train, \texttt{15\%} validation, \texttt{15\%} test respectively. The dataset exhibits an imbalanced class distribution, with some food categories and dishes having a higher number of samples, while others are underrepresented. For example, popular dishes like \textit{masala dosa} and \textit{biryani} have more images, whereas dishes like \textit{neer dosa} and \textit{chikki} have fewer samples. The average number of samples per class varies, and this imbalance requires techniques like data augmentation for better model performance. Figure \ref{fig:statistics} shows the number of images per food dish from the dataset.

\begin{figure}[h]
    \centering
    \begin{subfigure}{0.45\linewidth}
        \centering
    	\includegraphics[width=\linewidth]{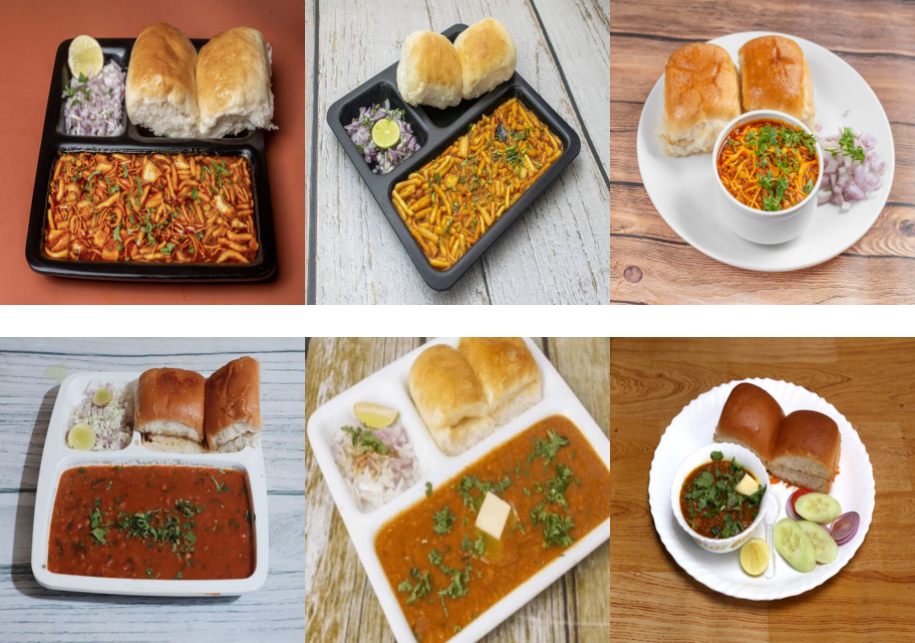}     
    	\caption{Similarities \textit{misal pav} and \textit{pav bhaji}}
    	\label{fig:similar-zero}
    \end{subfigure}
    \hfill
    \begin{subfigure}{0.45\linewidth}
        \centering
    	\includegraphics[width=\linewidth]{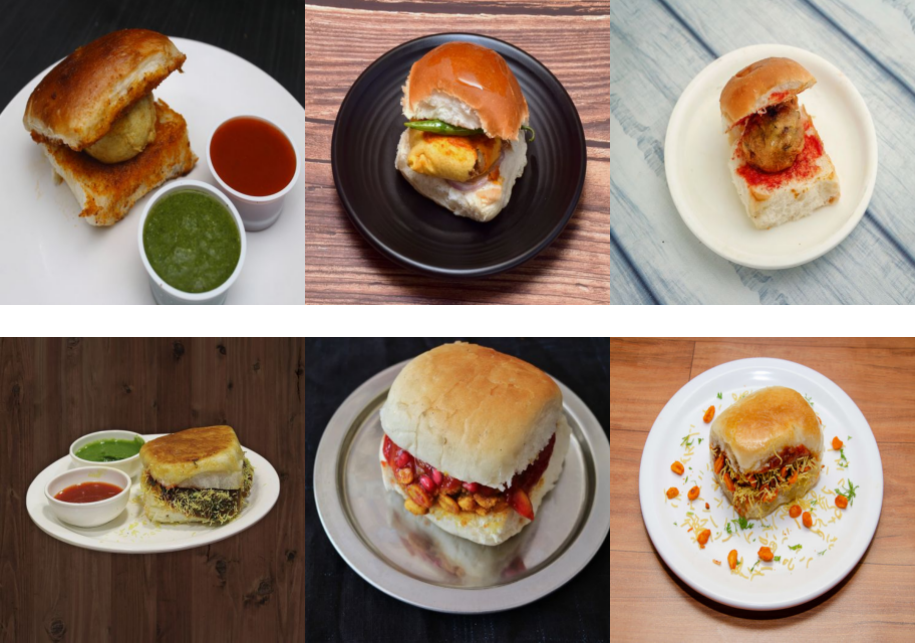}
    	\caption{Similarities of \textit{vada pav}  and \textit{dabeli}}
    	\label{fig:similar-one}
    \end{subfigure}
    \caption{Similarities between food dishes}
    \label{fig:similar-representations}
\end{figure}

\begin{figure}[h]
	\centering
	\includegraphics[width=0.75\linewidth]{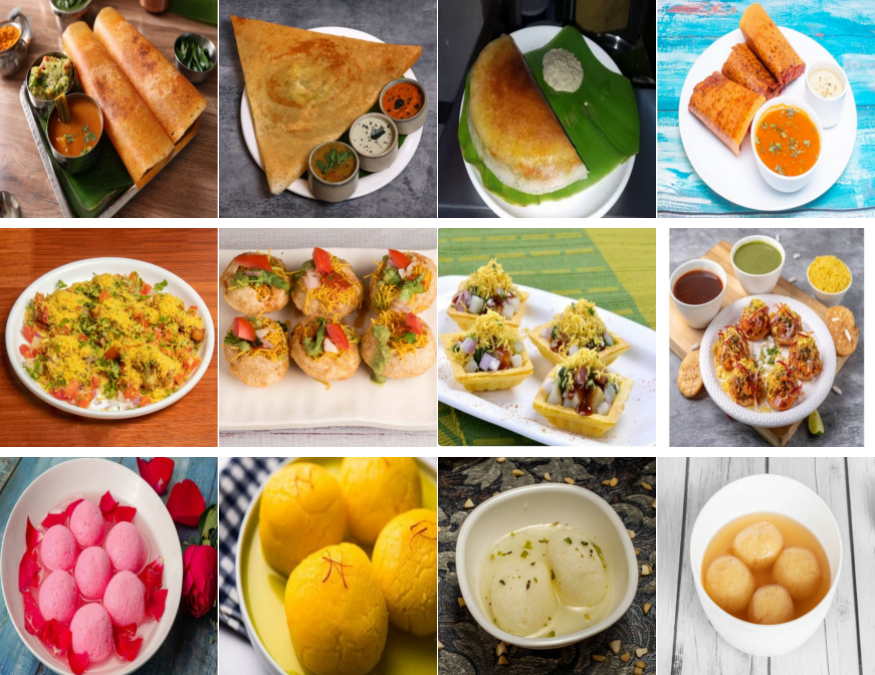}
	\caption{Different visual representations of single food dish}
	\label{fig:varied-representations}
\end{figure}

Figure \ref{fig:varied-representations} showcases the diverse visual representations within a single food dish. It is crucial to include these variations when creating datasets for food recognition as they enable more accurate and generalizable models. Figure \ref{fig:similar-representations} illustrates striking visual similarities between two distinct food dishes. These examples aid in training models to achieve precise classification accuracy and make them resilient to misclassification errors.

\section{Experiments}
\label{sec:experi}

\subsection{Experimental Setup}
\label{subsec:experi-setup}

In this section, we outline the image classification baselines selected for comparative analysis and the ideas driving their selection along with the setup for running the experimental analysis. Our emphasis was predominantly on leveraging pre-trained architectures like Convolutional Neural Networks (CNNs) and Transformers that are known for their proficiency in image classification problems. We opted for following models: 

\begin{table*}
  \centering
  \renewcommand{\arraystretch}{1.25}
  \begin{tabular}{ccc}
    \hline
    \textbf{Model} & \textbf{Total Params} & \textbf{Trainable Params} \\
    \hline
    ResNet-152 & 58,246,258 & 102,450 \\
    EfficientNet-V2-S & 20,241,538 & 64,050 \\
    ViT-B-16 & 85,837,106 & 38,450 \\\hline
    ConvNeXT-S & 49,493,138 & 38,450 \\\hline
  \end{tabular}
  \caption{Comparison of image classification model parameters}
  \label{tab:comparison-params}
\end{table*}

\begin{itemize}
  \item \textbf{Residual Networks (ResNet)}: ResNet \cite{Kaiming2016} have proven to be a strong baseline for comparison in computer vision research. We used this deep convolutional neural network architecture known for its residual learning approach and leveraged the pre-trained weights of \textit{ResNet-152}, which capture essential generic image features from the extensive ImageNet dataset and provide a robust starting point. It facilitates more efficient learning of task specific features with an unbalanced food dataset. Additionally, it might not be an optimal choice for all classification tasks due to potential overfitting.
  \item \textbf{EfficientNet}: EfficientNetV2 \cite{Tan2019,Tan2021} is a lightweight and efficient convolutional neural network architecture known for achieving high accuracy while requiring fewer parameters and computational resources when compared to other models. It utilizes a compound scaling method, uniformly scaling network depth, width, and resolution to maintain efficiency. EfficientNet has proved to be an industry accepted solution for small to medium scale image classification tasks. We used \textit{EfficientNet-V2-S} variant pre-trained on ImageNet for learning specific features from our dataset quickly, serving as a strong starting point for further improvement.
  \item \textbf{Vision Transformer (ViT)}: Vision Transformers \cite{Dosovitskiy2021} has demonstrated high accuracy on various benchmarks, making it a valuable benchmark for comparison. It doesn’t rely on pre-defined assumptions about image features, potentially leading to better generalization, and requires fewer resources when compared to CNNs of similar accuracy. We chose \textit{ViT-B-16} or \textit{ViT-Base} variant pre-trained on ImageNet as a competitive baseline for comparison, allowing us to benchmark and validate its effectiveness.
  \item \textbf{ConvNeXT}: ConvNext \cite{Liu2022} utilizes standard convolutional modules without relying on self-attention mechanisms as in transformers, leading to a simple and more interpretable architecture. It has achieved competitive results on various image classification tasks, proving to be a strong baseline for comparison. We leveraged \textit{ConvNeXT-Small} or \textit{ConvNeXT-S} pre-trained on ImageNet as a baseline as it contains comparable model parameters. ConvNeXT offered a modular design, depthwise separable convolutions, and a residual-like set aggregation block, which gave a balance between accuracy and efficiency. These chosen baselines offer a diverse set of architectural choices and complexity levels, providing a comprehensive comparison for evaluating the performance of our proposed approach on the food dataset.
\end{itemize}

These chosen baselines offer a diverse set of architectural choices and complexity levels, providing a comprehensive comparison for evaluating the performance of our proposed approach on the food dataset. The experiment encompasses of fine-tuning a pre-trained model, which involves selectively immobilizing the weights of earlier layers while focusing training efforts solely on the final layers customized for the specific classification task. Optimization of the model is facilitated through the \texttt{Adam} optimizer, leveraging a learning rate parameter set at \texttt{0.001}. The loss function utilized is cross-entropy, a widely acknowledged metric in classification tasks. Training unfolds over a duration of \texttt{50} epochs, with a batch size of \texttt{64}, to iteratively refine the model’s performance. Our setup defines a transformation for pre-processing images. This transformation resizes images and then extracts a central square matching the model’s expected size as shown in \ref{tab:comparison-results} It ensures consistency and avoids exceeding model limitations. Pixel values are normalized using values from the ImageNet dataset. It involves subtracting the mean and dividing by the standard deviation for each color channel (\texttt{red, green, blue}). This process effectively removes common variations in pixel intensity across images, leading to improved model performance and convergence. Finally, for smooth rescaling during the resizing process we use bilinear interpolation.

\subsection{Results}
\label{subsec:experi-results}

\begin{figure}
    \centering
    \begin{subfigure}{0.45\linewidth}
        \centering
    	\includegraphics[width=\linewidth]{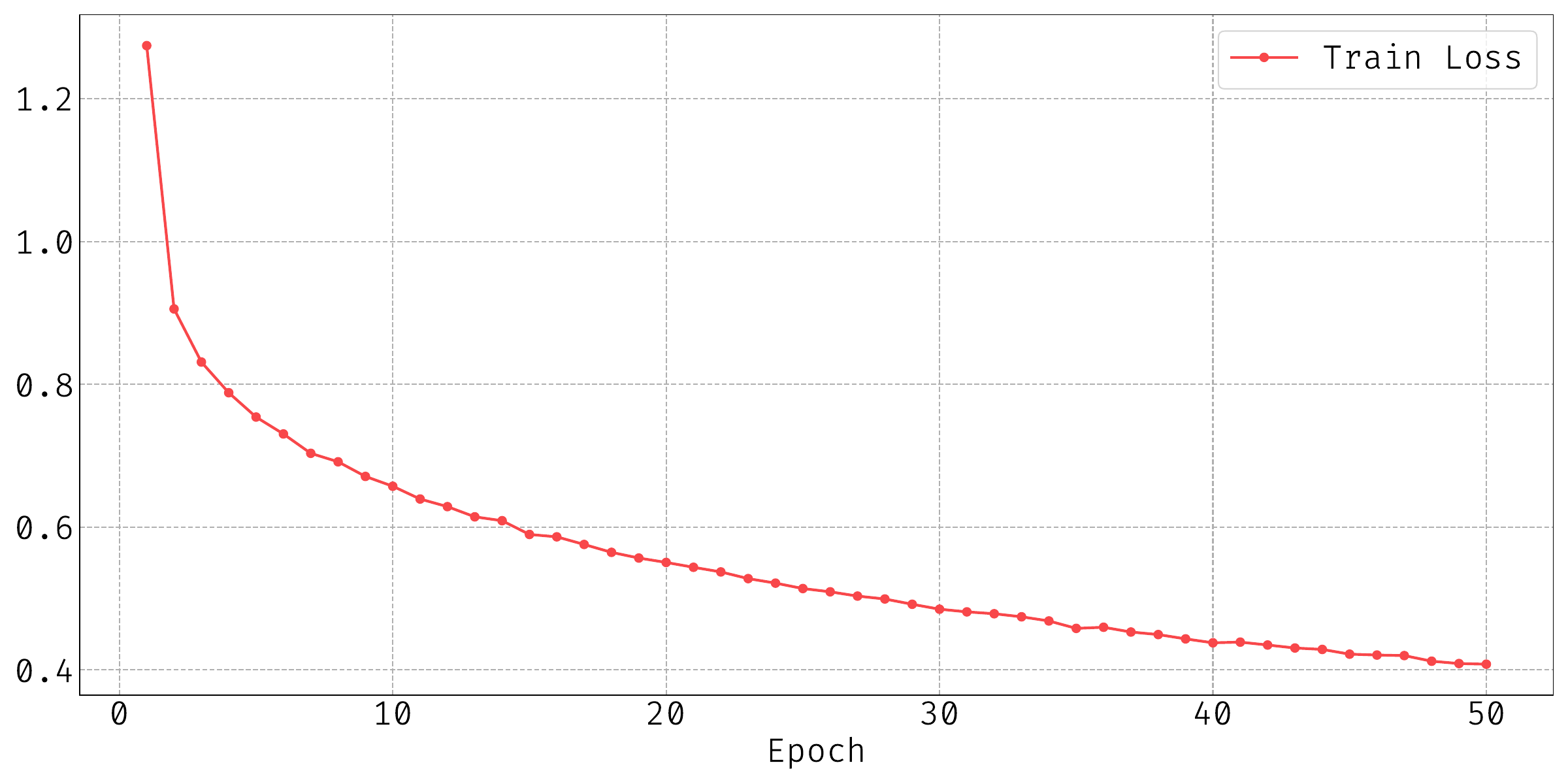}     
    	\caption{ResNet-152}
    	\label{fig:curve-resnet}
    \end{subfigure}
    \hfill
    \begin{subfigure}{0.45\linewidth}
        \centering
    	\includegraphics[width=\linewidth]{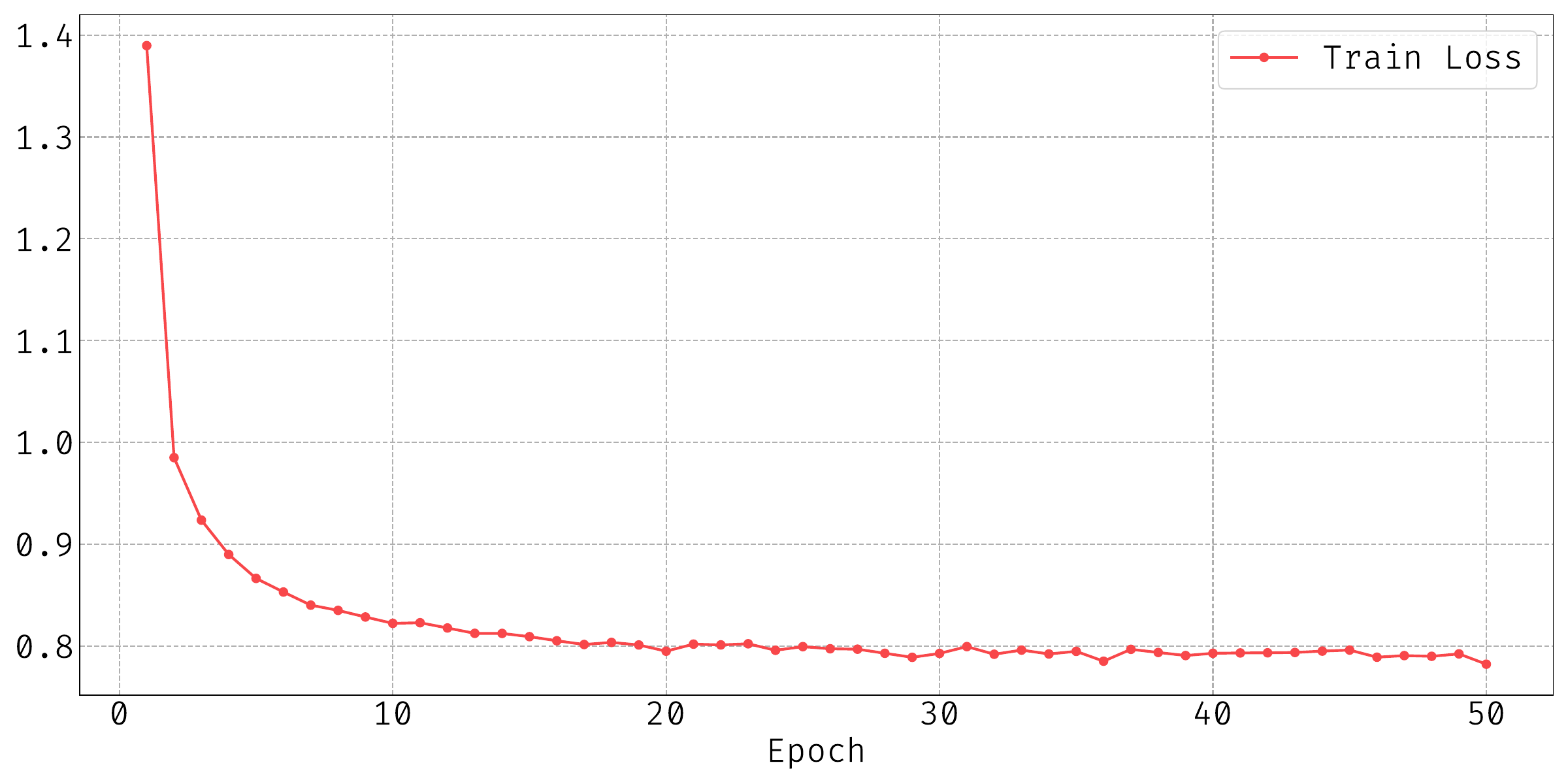}
    	\caption{EfficientNet-V2-S}
    	\label{fig:curve-efficientnet}
    \end{subfigure}
    \\[1.5em]
    \begin{subfigure}{0.45\linewidth}
        \centering
    	\includegraphics[width=\linewidth]{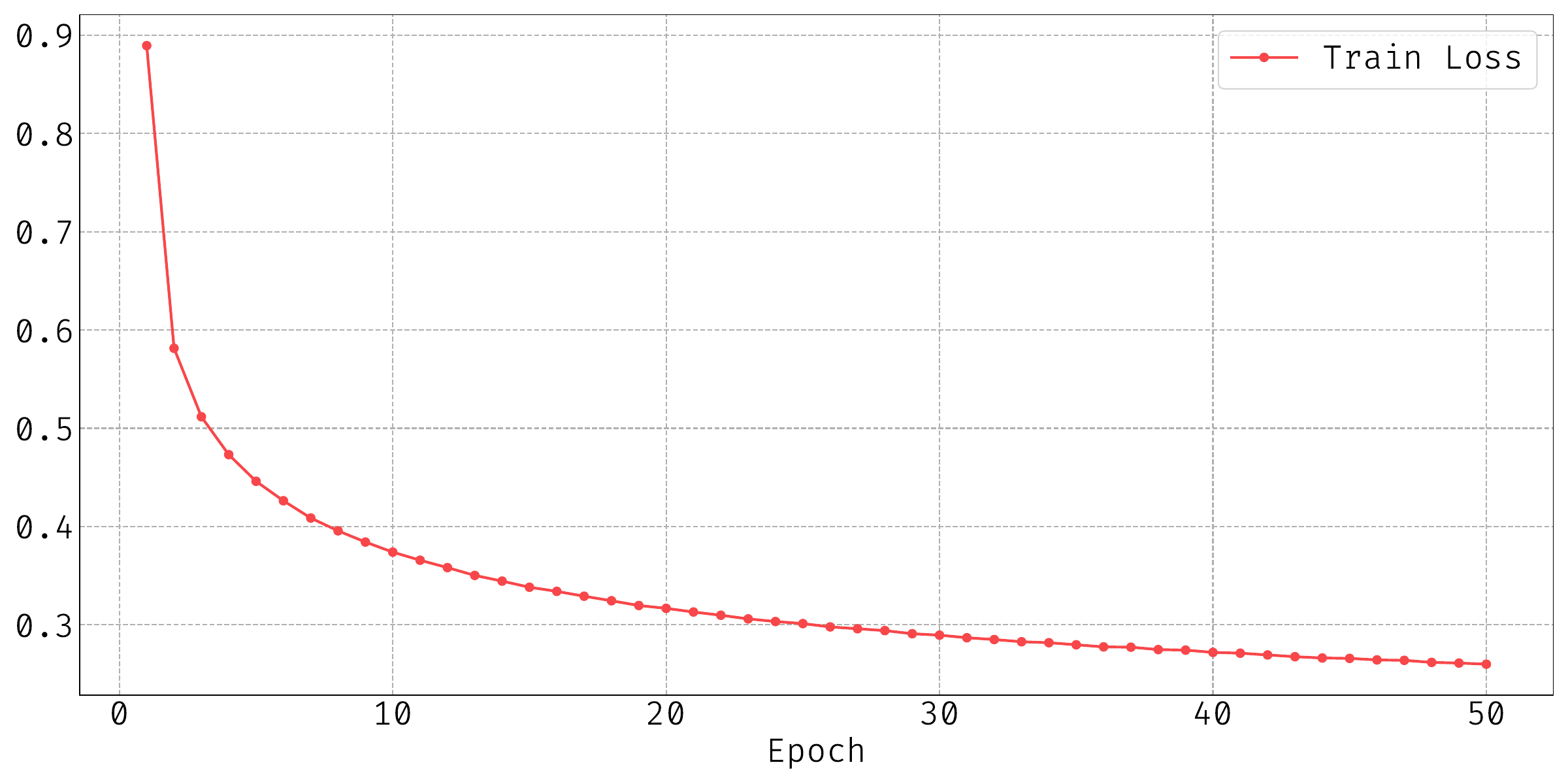}     
    	\caption{ViT-B-16}
    	\label{fig:curve-vit}
    \end{subfigure}
    \hfill
    \begin{subfigure}{0.45\linewidth}
        \centering
    	\includegraphics[width=\linewidth]{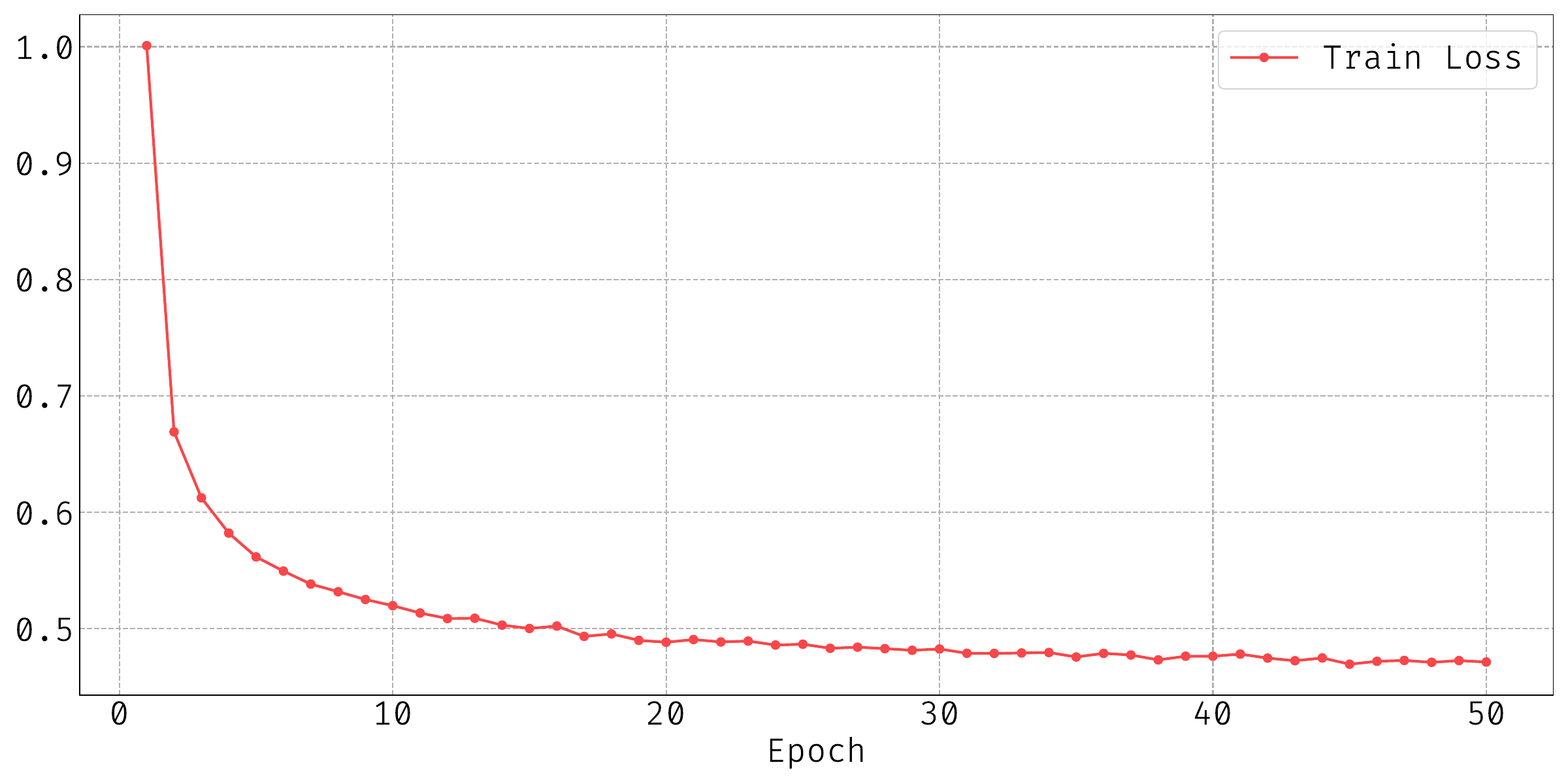}
    	\caption{ConvNeXT-S}
    	\label{fig:curve-convnext}
    \end{subfigure}
    \caption{Training loss curves for baseline models}
    \label{fig:list-curves}
\end{figure}

Out experimental analysis included comparing four state-of-the-art (SOTA) models: \textit{ResNet, EfficientNet, ViT, and ConvNeXT}. We evaluate the performance of each model based on loss curves and top-1 and top-5 accuracy. Figure \ref{fig:list-curves} shows the loss curves for the proposed method and each SOTA model during training. All models achieved convergence within \texttt{50} epochs. ViT reached the lowest final loss of \texttt{0.2}, followed by ResNet with a loss of \texttt{0.3} and ConvNeXT with a loss of \texttt{0.4}, while EfficientNet showed a slightly slower and less stable convergence behavior, reaching a final loss of \texttt{0.7}. Table \ref{tab:comparison-results} presents the top-1 and top-5 accuracy results for all models on the Khana dataset. \textit{ConvNeXT-S} model achieved the highest top-1 accuracy (\texttt{86.72\%}) and top-5 accuracy (\texttt{97.58\%}), outperforming the SOTA models with the lowest margin of \texttt{1.4\%} and \texttt{0.4\%}, for top-1 and top-5 accuracy, respectively.

\begin{table*}[h]
  \centering
  \renewcommand{\arraystretch}{1.25}
  \begin{tabular}{cccc}
    \hline
    \textbf{Model} & \textbf{Crop Size/Resize Size} & \textbf{Top-1 Accuracy} & \textbf{Top-5 Accuracy} \\
    \hline
    ResNet-152 & 224/232 & 81.00 & 95.37 \\
    EfficientNet-V2-S & 384/384 & 80.47 & 95.52 \\
    ViT-B-16 & 224/256 & 85.34 & 97.15 \\\hline
    \textbf{ConvNeXT-S} & \textbf{224/230} & \textbf{86.72} & \textbf{97.58} \\\hline
  \end{tabular}
  \caption{Comparison of top-1 and top-5 accuracy for baselines}
  \label{tab:comparison-results}
\end{table*}

\section{Limitations}

The dataset suffers from class imbalance, as evident from Figure \ref{fig:statistics}, with several popular food categories overrepresented and very niche food items underrepresented that leads to potential bias. Standard evaluation metrics do not capture fine-grained distinctions between food categories, and preprocessing can further influence the results. The dataset is promising for our given scope, but using it in applications to other contexts or datasets with different cultural and environmental conditions may lead to variability in results.

Exploring beyond this study's scope, we identify that adding more images for underrepresented food categories, expanding to new cuisines, and improving annotations are crucial for making models trained on this dataset more accurate for real-world applications. Given that the existing dataset covers a wide range of categories, we can expand beyond the baselines with new image classification models. Given the rapid growth in usage of multi-modal LLMs, researchers can explore querying over images and conducting qualitative comparisons of embeddings for several multi-modal LLMs as well.

\section*{Acknowledgments}

We want to thank Arpita Rane, Akshay Pithadiya, John D’souza, and Jane D’souza for their help in annotating a small sample of images during the development of an initial version of this dataset. 

\bibliographystyle{plainnat}
\bibliography{references}
\end{document}